\documentclass[10pt,twocolumn,letterpaper]{article}

\usepackage{cvpr}              
\definecolor{cvprblue}{rgb}{0.21,0.49,0.74}
\usepackage[pagebackref,breaklinks,colorlinks,allcolors=cvprblue]{hyperref}
\usepackage{bm}
\usepackage{enumitem}
\usepackage{multirow} 
\newtheorem{definition}{Definition}
\newtheorem{proposition}{Proposition}
\usepackage{tcolorbox}
\usepackage{pifont}
\usepackage{longtable}
\usepackage{makecell}
\usepackage{caption}
\usepackage{colortbl}
\usepackage{algorithm}
\usepackage{algpseudocode}
\tcbset{
    colback=cyan!5!white,    %
    colframe=red!25,  %
    coltitle=black, %
    boxrule=0.5pt,
    arc=2pt,
    left=6pt,
    right=6pt,
    top=6pt,
    bottom=4pt
}
\definecolor{ourpurple}{HTML}{CB297B}
\definecolor{ourlightpurple}{HTML}{FDF7FA}
\definecolor{ourblue}{HTML}{0076BA}
\definecolor{ourlightblue}{HTML}{F5FAFE}
\definecolor{ourmiddle}{HTML}{66509B}
\definecolor{ourlightmiddle}{HTML}{F7F6F9}
\hypersetup{colorlinks=true, allcolors=ourblue}

\def\paperID{} 
\def\confName{CVPR}
\def\confYear{2026}

\title{GARDO: Reinforcing Diffusion Models without Reward Hacking} 

\author{
\textbf{Haoran He}$^{1,2}$ ~ \textbf{Yuxiao Ye}$^{1}$ ~ \textbf{Jie Liu}$^{3}$ ~  \textbf{Jiajun Liang}$^{2}$ ~ \textbf{Zhiyong Wang}$^{4}$ \\
\textbf{Ziyang Yuan}$^{2}$ ~ \textbf{Xintao Wang}$^{2}$ ~ \textbf{Hangyu Mao}$^{2}$ ~ \textbf{Pengfei Wan}$^{2}$ ~ \textbf{Ling Pan}$^{1}$\\
{\normalsize{$^{1}$Hong Kong University of Science and Technology $^{2}$Kuaishou Technology $^{3}$CUHK MMLab $^{4}$The University of Edinburgh }} \\
\normalsize{\texttt{haoran.he@connect.ust.hk}}
}
\begin{document}

\maketitle

\begin{abstract}
Fine-tuning diffusion models via online reinforcement learning (RL) has shown great potential for enhancing text-to-image alignment. However, since precisely specifying a ground-truth objective for visual tasks remains challenging, the models are often optimized using a proxy reward that only partially captures the true goal. This mismatch often leads to reward hacking, where proxy scores increase while real image quality deteriorates and generation diversity collapses. While common solutions add regularization against the reference policy to prevent reward hacking, they compromise sample efficiency and impede the exploration of novel, high-reward regions, as the reference policy is usually sub-optimal. To address the competing demands of sample efficiency, effective exploration, and mitigation of reward hacking, we propose \textbf{G}ated and \textbf{A}daptive \textbf{R}egularization with \textbf{D}iversity-aware \textbf{O}ptimization (\textbf{GARDO}), a versatile framework compatible with various RL algorithms. Our key insight is that regularization need not be applied universally; instead, it is highly effective to selectively penalize a subset of samples that exhibit high uncertainty. To address the exploration challenge, GARDO introduces an adaptive regularization mechanism wherein the reference model is periodically updated to match the capabilities of the online policy, ensuring a relevant regularization target. To address the mode collapse issue in RL, GARDO amplifies the rewards for high-quality samples that also exhibit high diversity, encouraging mode coverage without destabilizing the optimization process. Extensive experiments across diverse proxy rewards and hold-out unseen metrics consistently show that GARDO mitigates reward hacking and enhances generation diversity without sacrificing sample efficiency or exploration, highlighting its effectiveness and robustness. Our project is available at \url{https://tinnerhrhe.github.io/gardo_project}

\end{abstract}    

\section{Introduction}

\label{sec:intro}

Denoising diffusion and flow models trained on large-scale datasets have found great success in text-to-image generation tasks, exhibiting unprecedented capabilities in visual quality~\citep{rombach2022high,flux2024,wu2025qwenimagetechnicalreport,xie2025sana}. While supervised pre-training provides these models a strong prior, it is often insufficient for ensuring alignment with human preferences~\citep{ouyang2022training,Uehara2024understanding}. Reinforcement Learning (RL) has emerged as a predominant paradigm to address this gap, either employing online policy-gradient approaches~\citep{DDPO,fan2023dpok,liu2025flowgrpo,gupta2025simple,div-rl,xue2025dancegrpo} or direct reward backpropagation~\citep{clark2024directly,xu2023imagereward}. These methods typically assume a reward model $r(x)$ that captures human preferences, and the training objective is to maximize the rewards over generated samples, i.e., $\mathbb{E}_{x\sim\pi}[r(x)]$. Consequently, the accuracy of $r(x)$ plays a critical role in fine-tuning performance. However, a key challenge arises from the nature of reward specification in vision. In contrast to the language domain, where rewards are often verifiable (e.g., the correctness of mathematical solutions or compiled code), rewards for visual tasks are typically more complex. Such rewards typically fall into two categories, both of which are imperfect proxies for genuine human preference: 1) \emph{Model-based rewards} such as ImageReward~\citep{xu2023imagereward}, UnifiedReward~\citep{unifiedreward}, and HPSv3~\citep{Ma2025HPSv3TW} are trained on finite human preference datasets to approximate a
ground-truth “genuine” reward, indicating that they are accurate only within their training data
distribution; 2) \emph{Rule-based rewards} such as object detection and text rendering (i.e., OCR)~\citep{chen2023textdiffuser} are limited to evaluating specific attributes, and thus fail to capture the overall qualities of generated samples. These limitations expose a significant distribution shift~\citep{wiles2021fine} vulnerability: fine-tuning methods can rapidly over-optimize the proxy reward by generating images that fall outside the trusted support of the reward model. This often yields spurious reward signals and thereby leads to \emph{reward hacking}~\citep{skalse2022defining}, where proxy rewards increase while the actual quality of the images degrades.

To address this challenge, prior work has employed KL-based regularization during fine-tuning to mitigate over-optimization on these spurious reward signals~\citep{uehara2024fine,fan2023dpok,liu2025flowgrpo}. However, as the online policy improves, the divergence from the static reference model can cause the KL penalty to dominate the RL loss. This often leads to diminishing policy updates, thereby impeding sample efficiency. Moreover, by design, the regularization constrains the online policy to remain in proximity to the reference model, which is often suboptimal. This can stifle effective exploration and prevent the discovery of emerging behaviors that are absent from the reference model.

\textit{Can we prevent reward hacking without compromising sample efficiency and effective exploration?} In this paper, we propose \textbf{G}ated and \textbf{A}daptive \textbf{R}egularization with \textbf{D}iversity-aware \textbf{O}ptimization (\textbf{GARDO}), a novel framework designed to overcome these limitations. Our gated regularization mechanism is motivated by the core insight: \textbf{KL penalty is not universally required}; Theoretically, only samples assigned with spurious rewards need regularization to prevent hacking. Accordingly, our gated mechanism applies the KL penalty selectively, targeting only a small subset (e.g., $\approx$10\%) of samples within each batch that exhibit the highest reward uncertainty~\citep{bai2022pessimistic}. We quantify this uncertainty by measuring the disagreement among an ensemble of reward functions, which serves as an effective proxy for the trustworthiness of the reward signal. High disagreement signifies that a generated sample likely falls into an out-of-distribution region where the reward models are extrapolating unreliably. To further accelerate training and sustain exploration, GARDO incorporates an adaptive regularization target. We periodically update the reference model with a recent snapshot of the online policy, facilitating continual improvement while retaining the stabilizing benefits of KL regularization. Finally, to enhance mode coverage, we introduce a diversity-aware optimization strategy. This is achieved by carefully amplifying the advantage term for high-quality samples that also exhibit high diversity. This amplification is carefully calibrated to ensure it neither dominates nor reverses the sign of the original advantage, which encourages broader exploration without destabilizing the optimization.

Our contributions are summarized as follows: (\romannumeral1) We provide a systematic analysis of the reward hacking phenomenon in RL-based fine-tuning for text-to-image alignment, identifying the core limitations of existing regularization techniques. (\romannumeral2) We make a novel innovation on traditional KL regularization. First, a gated regularization method selectively applies penalties only to high-uncertainty samples, thereby allowing the majority of samples to be optimized freely towards high-reward regions. Second, an adaptive regularization target facilitates continual improvement and sustained exploration by dynamically updating the reference anchor. (\romannumeral3) We propose a robust and generalist diversity-aware approach to improve generation diversity and mode coverage for diffusion RL. (\romannumeral4) Extensive experiments on multiple tasks across different proxies and unseen metrics demonstrate both the efficiency and effectiveness of our proposed method. Our work demonstrates that it is possible to successfully balance the competing demands of sample efficiency, exploration, and diversity, all while robustly mitigating reward hacking during the fine-tuning of image generative models.

\section{Related Works}
\label{sec:related_work}

\textbf{Fine-Tuning Diffusion Models via Rewards.}
Recent research has increasingly focused on fine-tuning pre-trained diffusion models using reward signals derived from human feedback. A reward model (RM) is trained on a finite dataset, learning to assign a scalar score that approximates true human preference~\cite{xu2023imagereward,liu2025improving, wu2025rewarddance,wang2025unified}. Typical methods include policy-gradient RL~\cite{black2023training, fan2024reinforcement, miao2024training}, direct preference optimization (DPO)~\cite{wallace2024diffusion, liu2025improving, yang2024using, liang2024step, yuan2024self, zhang2024onlinevpo,luo2025reinforcing}, and direct reward backpropagation~\cite{prabhudesai2023aligning, clark2023directly, xu2023imagereward}. Recently, Flow-GRPO~\cite{liu2025flowgrpo} and DanceGRPO~\cite{xue2025dancegrpo} have adapted GRPO for finetuning cutting-edge flow matching models, showing strong performance and inspiring subsequent research~\cite{li2025mixgrpo,he2025tempflow,wang2025pref,wang2025coefficients,fu2025dynamic,yu2025smart,zhou2025text}.

\noindent\textbf{Reward Hacking.}
Fine-tuning with either model-based or rule-based reward models for alignment of visual generation is highly susceptible to reward hacking~\cite{laidlaw2025correlated,skalse2022defining}. This issue arises because the reward model is an imperfect proxy for the true human preference. As a generative model optimizes against this proxy, it often discovers adversarial solutions that exploit the reward model's flaws, such as favoring extreme saturation or visual noise~\cite{clark2023directly, liu2025improving}, to achieve a high score, despite failing to satisfy human intent. To address this, standard methods like DPOK~\cite{fan2023dpok} and Flow-GRPO~\cite{liu2025flowgrpo} employ KL regularization to constrain the policy update. While this can prevent the most egregious forms of reward hacking, it often sacrifices convergence speed and hinders effective exploration. Another line of work, exemplified by RewardDance~\cite{wu2025rewarddance}, focuses on improving the reward model itself through scaling; however, this does not eliminate the vulnerability to out-of-distribution samples. Concurrently, Pref-GRPO~\cite{wang2025pref} utilizes a pairwise preference model to provide a more robust reward signal and avoid illusory advantages. This approach, however, is not universally applicable as it is limited to preference-based reward models and incurs substantial computational costs for pairwise comparisons.

\section{Preliminaries}
\subsection{Denoising as an MDP}
Both diffusion models and flow models map the source distribution, often a standard Gaussian distribution, to a true data distribution $p_0$. As shown in previous works~\citep{kim2025inference,singh2024stochastic,he2025scaling}, these models can utilize an SDE-based
sampler during inference to restore from diffused data. Inspired by DDPO~\citep{DDPO}, we formulate the multi-step denoising process in flow and diffusion-based models as a Markov Decision Process (MDP), defined by a tuple $(\mathcal{S},\mathcal{A},\mathcal{R},\mathcal{P},\rho_0)$. At each denoising step $t$, the model receives a state $s_t\triangleq(c,t,x_t)$, and predicts the action $a_t\triangleq x_{t-1}$ based on the policy $\pi(a_t|s_t)\triangleq p_\theta(x_{t-1}|x_t,c)$. A non-zero reward $R(s_t,a_t)\triangleq r(x_0,c)$ is given only at the final step, where $R=0$ if $t\neq0$. The transition function $\mathcal{P}$ is deterministic. At the beginning of each episode, the initial state $s_T$ is sampled from the initial state distribution $\rho_0$. The goal is to learn a policy $\pi^*=\arg\max_\pi\mathbb{E}_{\rho_0,a_t \sim \pi(s_t)}\big[R(s_0,a_0)]$ by maximizing the expected cumulative reward $R$.

\subsection{RL on Diffusion and Flow Models}

Reinforcement Learning (RL)~\citep{DDPO,fan2023dpok} has been widely used for enhancing sample quality by steering outputs towards a desired reward function. Unlike earlier approaches based on PPO~\citep{schulman2017proximal} or REINFORCE~\citep{williams1992simple}, recent methods~\citep{liu2025flowgrpo,xue2025dancegrpo} have demonstrated greater success using GRPO~\citep{shao2024deepseekmath}. We take GRPO as the base RL algorithm throughout our paper. GRPO rollouts $\{x_0^i\}_{i=1}^G$ samples conditioned on the same input $c$, and estimates the advantage within each group:
\begin{equation}
\label{eq:adv_norm}
    A_t^i=\frac{R(x_0^i)-{\rm mean}(\{R(x_0^i)\}^G_{i=1})}{{\rm std}(\{R(x_0^i)\}^G_{i=1})}, \forall t.
\end{equation}
Then the following surrogate objective optimizes $\pi_\theta$:
\begin{equation}
\begin{aligned}J(\theta)&=\mathbb{E}_{x_0\sim\pi_{\theta_{\rm old}}} \big[\frac{1}{G}\sum_{i=1}^G\frac{1}{T}\sum\nolimits_{t=0}^{{T}}\big(\min\big({\rm r}^i_t A^i_t,\\&{\rm clip}({\rm r}^i_t, 1-\epsilon,1+\epsilon)A^i_t\big)-\beta D_{KL}(\pi_\theta|\pi_{\rm ref})\big)\big],
\end{aligned} 
\end{equation}
with ${\rm r}^i_t={\pi_{\theta}(a_t|s_t)}/{\pi_{\theta_{\rm old}}(a_t|s_t)}$ the importance sampling ratio, $\pi_{\theta_{\rm old}}$ the behavior policy to sample data, $\epsilon$ the clipping range of ${\rm r}_t$, and $D_{KL}$ the KL regularization term.

\section{Method}

We first present the reward hacking issues in image generation as a motivated example. Subsequently, we provide a theoretical analysis, and propose practical methods to address the problem. Finally, we provide a didactic example for empirical validation to illustrate our method clearly. 

\subsection{Reward Hacking in RL-Based Fine-tuning}

\begin{figure}
    \centering
    \includegraphics[width=1\linewidth]{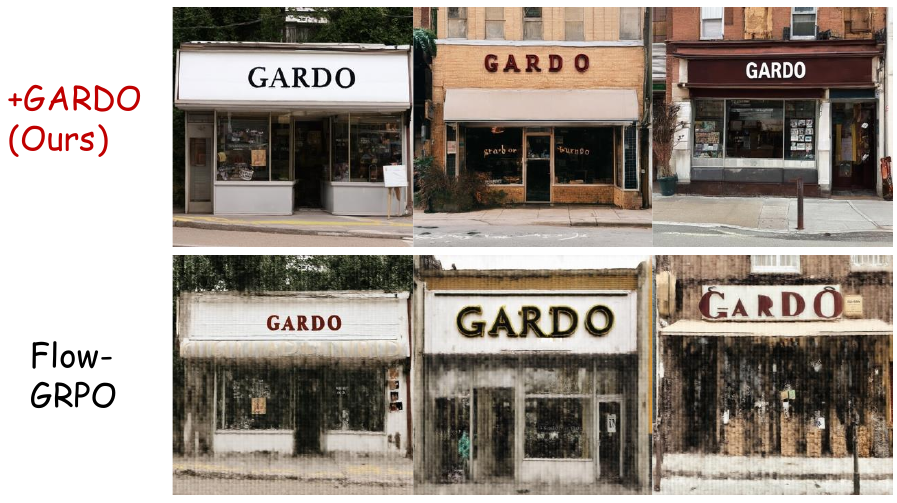}
    \vspace{-1.em}
    \caption{Setting OCR as the proxy reward, vanilla RL methods like Flow-GRPO exploit the OCR reward at the cost of losing real image quality, leading to reward hacking. It generates unrealistic, noisy images with blurry backgrounds and visual artifacts. In contrast, our method maintains better image quality and diversity. Prompt:``A storefront with `GARDO' written on it".}
    \vspace{-1.em}
    \label{fig:hack_comp}
\end{figure}
\begin{figure*}[t]
    \centering
    \includegraphics[width=1\linewidth]{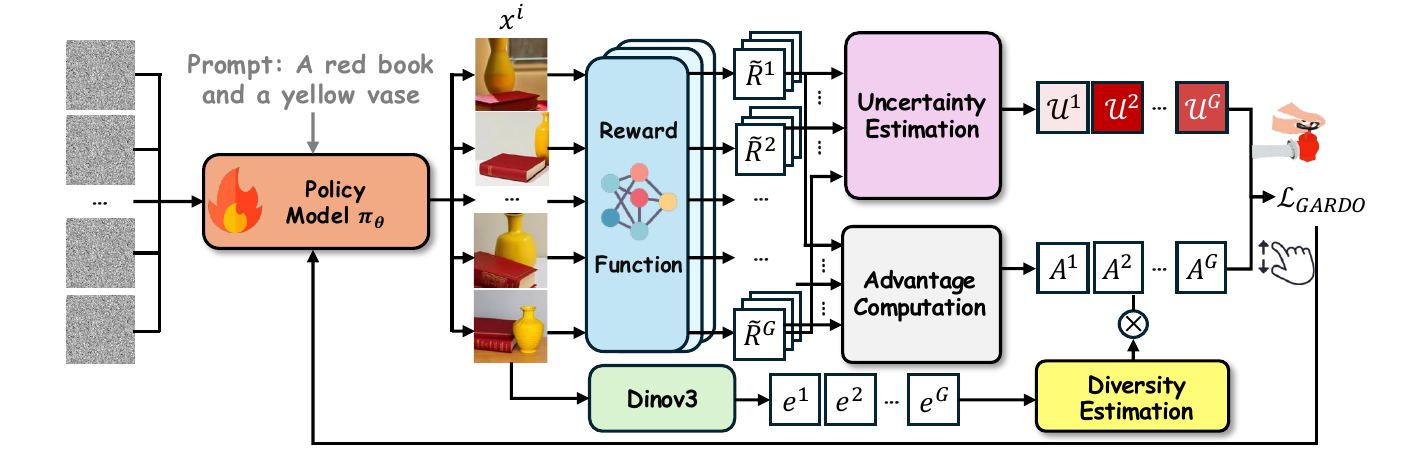}
    \vspace{-1.em}
    \caption{Overview of GARDO. GARDO introduces an uncertainty-driven, gated KL mechanism to control the proportion of regularization, avoiding unnecessary penalties. Our proposed diversity-aware advantage shaping effectively encourages exploration of novel states. }
    \label{fig:overview}
    \vspace{-1.em}
\end{figure*}

We investigate RL in the context of fine-tuning text-to-image models. Here, the reward function is often a neural network trained on finite preference data or a rule-based metric, which serves as just a proxy for the true reward. Misalignment between the two objectives can lead to \emph{reward hacking}: a learned policy performs well according to the proxy reward but not according to the true reward~\citep{pan2022the}. 

\begin{definition}
\label{def:hack}
    {\rm (Reward hacking~\citep{laidlaw2025correlated}).} Suppose $\tilde{R}$ is a proxy reward used during RL fine-tuning, and $R$ denotes the true reward. $\tilde{R}$ is a hackable proxy with respect to $\pi_{ref}$ s.t.
    \begin{equation}
        J(\pi,R)<J(\pi_{\rm ref},R)\ {\rm for\ some}\ \pi\in\arg\max_\pi J(\pi,\tilde{R}), 
    \end{equation}
    where $J(\pi,R)=\mathbb{E}_\pi[\sum_tR(s_t,a_t)]$ is the RL objective, and $\pi_{\rm ref}$ is the reference policy.
\end{definition}

As demonstrated in Figure~\ref{fig:hack_comp}, standard methods like Flow-GRPO~\citep{liu2025flowgrpo} maximize the proxy score (i.e., OCR) at the expense of perceptual quality, failing to align with the true human preferences. Previous methods~\citep{uehara2024fine} address this issue by incorporating a KL divergence regularization term, 

\begin{equation}
\label{eq:kl_obj}J_\beta(\pi_\theta)=\mathbb{E}_{\pi_{\theta}}\big[\sum_t\tilde{R}(s_t,a_t)\big]-\beta D_{\rm KL}(\pi_\theta|\pi_{\rm ref}),
\end{equation}
where $\beta$ controls the regularization strength. This kind of approach mitigates reward hacking (or reward over-optimization) by explicitly penalizing significant deviations of the online policy, $\pi$, from the reference policy, $\pi_{\rm ref}$. However, this strategy introduces two significant limitations. First, because the reference policy $\pi_{\rm ref}$ is typically suboptimal, the regularization loss can impede learning, leading to poor sample efficiency. Second, the constant penalty constrains $\pi_{\theta}$ to exploit the ``safe'' region around $\pi_{\rm ref}$, preventing it from discovering novel behaviors or solutions that are absent from the reference model. 

Ideally, for fine-tuning image generative models via RL, our goal is to reconcile the following competing objectives: (\romannumeral1) optimizing $\pi_\theta$ with high sample efficiency, (\romannumeral2) avoiding reward hacking and over-optimization, (\romannumeral3) encouraging exploration to high-reward, novel modes which may be missing from $\pi_{\rm ref}$,
and (\romannumeral4) preserving generation diversity. Conventional KL regularization is effective in accomplishing (\romannumeral2) and (\romannumeral4), but often at the expense of (\romannumeral1) and (\romannumeral3). This inherent trade-off motivates our work. Based on our theoretical findings, we propose \textbf{G}ated and \textbf{A}daptive \textbf{R}egularization with \textbf{D}iversity-aware \textbf{O}ptimization (\textbf{GARDO}), a novel framework that is simple, effective, and capable of satisfying all these objectives simultaneously. Overview of GARDO is illustrated in Fig.~\ref{fig:overview}, and we detail our method in the subsequent sections.

\subsection{Gated and Adaptive Regularization}
First, we investigate the reasons of reward hacking. As discussed in previous works~\citep{gx2025kl,wu2023practical,Uehara2024understanding}, the optimal solution to the KL-regularized objective in Eq~\eqref{eq:kl_obj} can be written as
\begin{equation}
\label{eq:kl_target}
    p^*(x)=\frac{1}{\mathcal{Z}}\pi_{\rm ref}(x)\exp\left(\frac{\tilde{R}(x)}{\beta}\right),
\end{equation}
where $\mathcal{Z}$ is an intractable normalization constant~\citep{rafailov2023direct}. Here, $p^*(x)$ is largely defined by both the reference model $\pi_{\rm ref}(x)$ and the proxy reward $\tilde{R}(x)$.
\begin{proposition} The probability ratio between any two samples, $x^1$ and $x^2$, under the optimal solution distribution defined in Eq.~\eqref{eq:kl_target} is given by the following closed form,
    \begin{equation}
    \frac{p^*(x^1)}{p^*(x^2)}=\frac{\pi_{\rm ref}(x^1)}{\pi_{\rm ref}(x^2)}\exp\left(\frac{\tilde{R}(x^1)-\tilde{R}(x^2)}{\beta}\right).
\end{equation}
\label{proposition1}
\end{proposition}
If $\pi_{\rm ref}(x^1)=\pi_{\rm ref}(x^2)$, then their probability difference is defined solely by the proxy reward, i.e., $p^*(x^1)/p^*(x^2)=\exp((\tilde{R}(x^1)-\tilde{R}(x^2))/\beta)$. Reward hacking arises when the proxy reward is misaligned with the true reward, $R$. For example, if $\tilde{R}(x^1)>\tilde{R}(x^2)$ while $R(x^1)<R(x^2)$, the model will incorrectly assign a higher sampling probability to the lower-quality sample, $x^1$. However, in other cases, such as `$x^1>x^2$' holds for both $\tilde{R}$ and $R$, the proxy is reliable and does not lead to reward hacking. Inspired by this finding, we have the following key insight: 

\begin{tcolorbox}
[title=\emph{\textbf{Takeaway 1}}]
    Regularization is not universally required. It is required only for samples with spurious proxy reward $\tilde{R}$.
\end{tcolorbox}
The application of KL regularization should be conditional on the alignment between proxy and true rewards. Universally penalizing all samples introduces unnecessary regularization signals, thereby impeding convergence. 
\begin{figure*}[!t]
    \centering
    \includegraphics[width=1\linewidth]{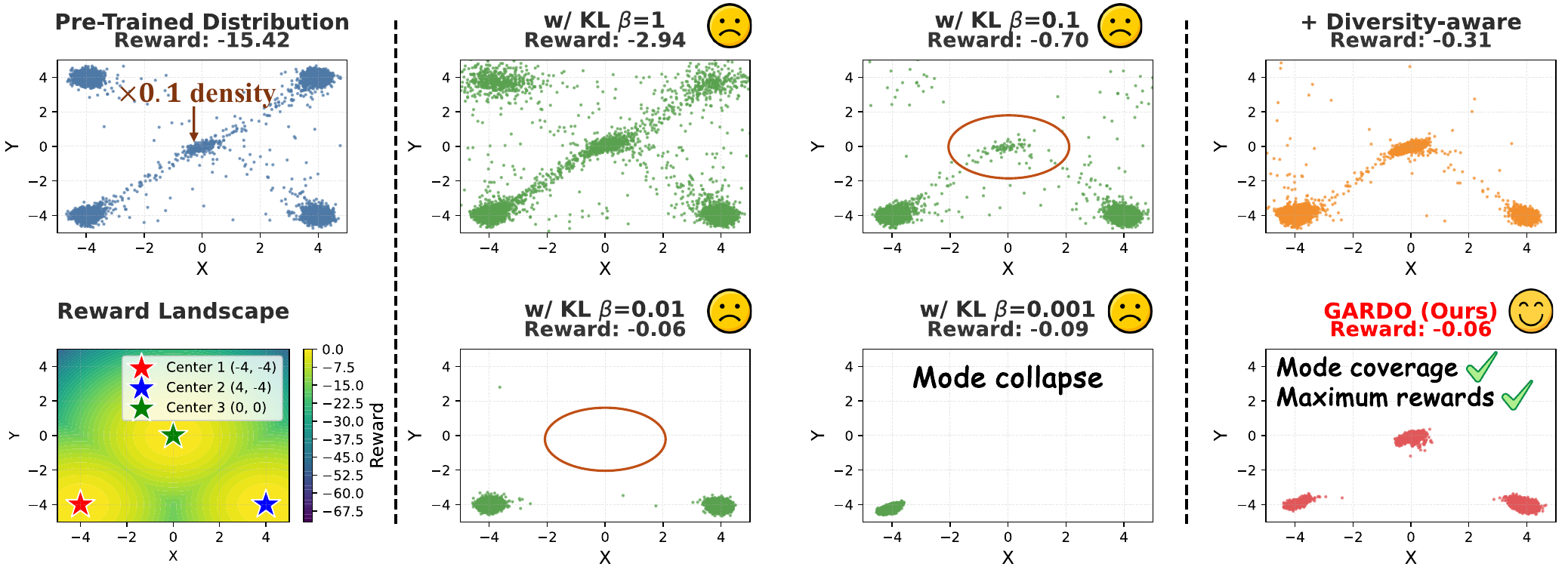}
    \vspace{-1em}
    \caption{We train a diffusion model with $3$-layer MLP on Gaussian mixtures (pre-trained distribution), with the goal to capture multimodal high-reward clusters as shown in the reward landscape. The vanilla RL method (DDPO~\citep{DDPO}) with a large KL coefficient $\beta$ is overly constrained and fails to increase rewards. Conversely, a small $\beta$ incurs severe mode collapse. Our proposed diversity-aware optimization, when applied alone, successfully captures the multimodal modes, including the central cluster with the lowest probability density in the reference policy $\pi_{\rm ref}$. Our full GARDO framework simultaneously achieves maximum reward and discovers all high-reward clusters.}
    \vspace{-1em}
    \label{fig:toy-case}
\end{figure*}

\noindent\textbf{Gated KL Mechanism.} KL is gated for selective samples. We propose an uncertainty-driven approach to select samples for regularization during the training process~\citep{an2021uncertainty}. Only samples that exhibit high uncertainty $\mathcal{U}$ will be penalized, where $\mathcal{U}$ reflects the trustworthiness of the proxy reward~\citep{clements2019estimating}. Our approach for quantifying uncertainty is inspired by prior work in reinforcement learning~\citep{bai2022pessimistic}, where the uncertainty is often estimated using the disagreement within an ensemble learned $K$ bootstrapped Q-functions $Q^k$~\citep{osband2016deep}, i.e., $\mathcal{U}(s_t,a_t):={\rm Std}(Q^k(s_t,a_t))$. In the denoising MDP of image generation, considering the proxy $\tilde{R}$ is a function of the final state $x_0$, we can simplify the uncertainty quantification as $\mathcal{U}(s_t):={\rm Std}(V^k(s_t))$. However, directly learning ensemble $K$ value functions from scratch is computationally prohibitive for large-scale generative models, given their vast state space and the complexity of the image generation process. To circumvent this challenge, we adopt a more practical approach that approximates the value function using readily available, pre-trained reward models~\citep{li2024derivative}. Instead of leveraging the deviation among ensemble value functions for estimating uncertainty, we propose a new uncertainty quantification approach:
\begin{equation}
\label{eq:uncertainty}
    \mathcal{U}(x^i):=w(\tilde{R}(x^i))-{\rm mean}(\{w(\hat{R}_n(x^i))\}_{n=1}^K),
\end{equation}
where $w(y^i)=\frac{1}{B}\sum_{j\neq i}\mathbb{I}(y^i>y^j)$ denotes the win rate within a batch of size $B$, and $\{\hat{R}_n\}_{n=1}^K$ are the auxiliary reward models. Under this formulation, a high uncertainty score arises when $w(\tilde{R})\gg{\rm mean}(\{w(\hat{R}_n)\}_{n=1}^K)$, effectively flagging samples with anomalously high proxy $\tilde{R}$ compared to the ensemble. We choose light-weight Aesthetic~\citep{schuhmann2022laion} and ImageReward~\citep{xu2023imagereward} as $\hat{R}$ throughout our paper (i.e., $K$=2), thereby the computation cost is negligible. Note that only $\tilde{R}$ is optimized during fine-tuning, while $\hat{R}$ only serves as a metric for estimating uncertainty. Surprisingly, a key empirical finding of our work is that this selective penalization is highly efficient. We find that applying the KL penalty to only a small subset of samples (e.g., approximately 10
\%) with the highest uncertainty is sufficient to prevent reward hacking.

\noindent\textbf{Adaptive KL Regularization.} While the gated KL mechanism penalizes only high-uncertainty samples, we observe that the sample efficiency still slows as training progresses. As indicated in Proposition~\ref{proposition1}, $\pi_{\text{ref}}$ also plays a pivotal role in determining the optimal distribution. If $\tilde{R}(x^1)=\tilde{R}(x^2)$, a failure mode arises if $\pi_{\text{ref}}(x^1)\ll\pi_{\text{ref}}(x^2)$, resulting $p^*(x^1)\ll p^*(x^2)$, which is not an expected behavior. We remark that a static reference model becomes increasingly sub-optimal, particularly at later training stages, which can affect the optimization a lot. To mitigate this limitation and facilitate sustained improvement, we propose an adaptive regularization objective. We periodically hard-resets the reference model $\pi_{\text{ref}}$ to the current policy at specific epochs, allowing it to remain updated. In particular, the reference model is updated whenever the KL divergence $D_{\rm KL}$ surpasses a pre-defined threshold $\epsilon_{\rm KL}$, or, failing that, after a maximum of $m$ gradient steps. Our proposed adaptive KL mechanism ensures the regularization target remains relevant, preventing the KL penalty from dominating the RL loss and halting exploration, thereby achieving sustained policy improvement. 

\begin{tcolorbox}[title=\emph{\textbf{Takeaway 2}}]
A static reference model inevitably becomes a constraint on RL optimization. Dynamically updating the reference model facilitates prolonged improvement.
\end{tcolorbox}

\subsection{Diversity-Aware Optimization}

As noted by \citet{liu2025flowgrpo}, a significant consequence of reward hacking is the reduced diversity. This issue is exacerbated by the intrinsically mode-seeking nature of reinforcement learning, which often struggles to capture multimodal distributions~\citep{kim2025testtime}. Enhancing sample diversity is therefore critical not only for preventing mode collapse but also for broadening the policy's exploration space. While our gated and adaptive regularization scheme effectively mitigates reward hacking without sacrificing sample efficiency, it does not explicitly promote generation diversity. To address this limitation, we introduce a diversity-aware optimization strategy that amplifies rewards for high-quality samples that also exhibit high diversity.

\noindent\textbf{Diversity-Aware Advantage Shaping.} The core idea is to reshape the advantage function by incorporating a diversity-based signal during policy optimization. Specifically, for a group of generated samples, $\{x_0^i\}_{i=1}^G\sim p_\theta(x_0|c)$, we first map each sample from the pixel space into a semantic feature space, obtaining feature embeddings $e^i$ for each clean image $x_0^i$. We employ DINOv3~\citep{simeoni2025dinov3} for the feature extraction, which serves as a powerful vision foundation model. A sample's diversity is then quantified by its isolation in the feature space. We define a diversity score, $d_i$, as the cosine distance to its nearest neighbor within the group, $\{e^i\}_{i=1}^G$. This diversity score is subsequently used to reshape the advantages, as illustrated as follows:
\begin{equation}
\label{eq:diversity}
\begin{aligned}
    &e_i={\rm Dinov3}(x_0^i),\ d_i=\min (\{1-\frac{e_i\cdot e_j}{|e_i||e_j|}\}_{j\neq i}),\\ &A_i^{\rm shaped}=A_i\cdot {\rm d}_i\ \ \text{if}\ A_i>0\ \text{else}\ A_i,\ \ \ i,j\sim[1,G].
\end{aligned}
\end{equation}
There are several key design principles for our proposed diversity-aware advantage reshaping: (1) We use a multiplicative re-weighting of the advantage term rather than an additive diversity bonus. This design circumvents the need for delicate hyperparameter tuning to balance the scales of the proxy reward and the diversity score, which could otherwise cause one signal to dominate the other. (2) The advantage shaping is applied only when a sample's advantage is positive. This is a critical constraint that ensures the model is rewarded only for generating samples that are both high-quality and novel. It explicitly prevents the model from generating aberrant or low-quality images simply to increase its diversity score. We summarize our method in the pseudocode in Alg.~\ref{alg:gardo}.

\begin{tcolorbox}[title=\emph{\textbf{Takeaway 3}}]
    Multiplicative advantage reshaping exclusively within positive samples enables robust diversity improvement.
\end{tcolorbox}

\begin{table*}[!t]
\caption{Results of GARDO and diverse baselines across both proxy rewards and diverse o.o.d. rewards. The proxy task is marked by the orange color.
}
\label{table:main_result}
\begin{center}
\resizebox{\linewidth}{!}{
\begin{tabular}{ccllllllll}
\toprule
 \multirow{2}{*}{\bf Method}& \multirow{2}{*}{\bf \#Step} & \multicolumn{2}{c}{\bf Trained Tasks}& \multicolumn{6}{c}{\bf Unseen Tasks}\\\cmidrule(lr){3-4} \cmidrule(r){5-10}
 &&\multirow{1}{*}{\bf GenEval}  & \multirow{1}{*}{\bf OCR} & \multicolumn{1}{l}{\bf Aesthetic} & \multicolumn{1}{l}{\bf PickScore} &\multicolumn{1}{l}{\bf ImgRwd} & \multicolumn{1}{l}{\bf ClipScore} &\multicolumn{1}{l}{\bf HPSv3} &\multicolumn{1}{l}{\bf Diversity}\\ 
\hline
\textcolor{gray}{SD3.5-M~\cite{rombach2022high}}  & - & \textcolor{gray}{0.63} & \textcolor{gray}{0.58} & \textcolor{gray}{5.07} & \textcolor{gray}{22.40}  & \textcolor{gray}{0.83} & \textcolor{gray}{28.2}& \textcolor{gray}{9.70}& \textcolor{gray}{21.84}\\\hline
\multicolumn{10}{c}{\emph{OCR Task}}\\
+GRPO ($\beta$=0)  & 600 &0.52 & \cellcolor[rgb]{1.0,0.9,0.8}\underline{0.93} &4.67  & 21.82 & 0.61 & 27.9&8.11&18.15\\
+GRPO ($\beta$=0) w/o std norm  &  600   & 0.57  &\cellcolor[rgb]{1.0,0.9,0.8}0.92  & 4.88 & 22.05 & 0.68 & 28.0&8.32&19.37\\
+GRPO ($\beta$=0.01) w/o std norm &600     &  \underline{0.64} & \cellcolor[rgb]{1.0,0.9,0.8}0.86 &\textbf{5.08}  &\textbf{22.45}  &\underline{0.90}  &28.6 &\textbf{9.89}&\underline{21.32}\\
+ GARDO (Ours) ($\beta$=0.04) w/o div & 600 & 0.63     & \cellcolor[rgb]{1.0,0.9,0.8}0.91&5.03&22.41&0.87 & 28.7 &9.22  & 19.89\\\hline
\rowcolor[rgb]{0.9,1.0,0.8}+GARDO (Ours) ($\beta$=0.04) &600  & \textbf{0.65}  &\cellcolor[rgb]{1.0,0.9,0.8}0.92   &\textbf{5.07}  &\textbf{22.41}  & \textbf{0.92} &\textbf{28.7}  & \underline{9.75}&\textbf{21.60}\\
\rowcolor[rgb]{0.9,1.0,0.8}+GARDO (Ours) ($\beta$=0.04) &1400  & 0.60  &\cellcolor[rgb]{1.0,0.9,0.8}\textbf{0.96}   &5.02  &22.31  & 0.87 &\textbf{28.7}  & 9.51&21.03\\
\midrule
\multicolumn{10}{c}{\emph{GenEval Task}}\\
+GRPO ($\beta$=0) &2000 & \cellcolor[rgb]{1.0,0.9,0.8}0.95& 0.60 &4.80 & 21.92 &0.73  &28.4 &6.73&15.6\\
+GRPO ($\beta$=0) w/o std norm  &  2000  & \cellcolor[rgb]{1.0,0.9,0.8}0.94  & 0.61 & 4.91 & 22.06 &0.79  &28.5 &6.91&15.91\\
+GRPO ($\beta$=0.01) w/o std norm & 2000   & \cellcolor[rgb]{1.0,0.9,0.8}0.81  & 0.64 & \textbf{5.15} & \textbf{22.5} &\textbf{0.97} &\underline{28.7}&\textbf{10.17}&\underline{21.73}\\
+ GARDO (Ours) ($\beta$=0.04) w/o div & 2000 &  \cellcolor[rgb]{1.0,0.9,0.8}0.95    &0.63 &5.01&22.1&0.90 & 28.6 & 8.61 &19.98 \\\hline
\rowcolor[rgb]{0.9,1.0,0.8}+GARDO (Ours) ($\beta$=0.04) & 2000 &  \cellcolor[rgb]{1.0,0.9,0.8}\textbf{0.95} & \textbf{0.68}  & \underline{5.09} & \underline{22.34} & \underline{0.95} & \textbf{29.4} &\underline{9.27} &\textbf{24.95}\\
\bottomrule
\end{tabular}
}
\end{center}
\end{table*}
\noindent\textbf{Empirical Validation.} As illustrated in Fig.~\ref{fig:toy-case}, we provide a didactic example to validate the superior efficacy of our method. We observe that only GARDO successfully captures all high-reward modes, reaching the maximum reward. The most notable outcome is GARDO's discovery of the central cluster, a mode with only $0.1\times$ probability density assigned by the reference model compared with other modes. This ``mode recovery" capacity underscores GARDO's potential for robust exploration, showing it can incentivize emerging behaviors that lie far from the pre-trained distribution.

\noindent\textbf{An Interesting Finding.} Beyond the above techniques, we find that simply removing the standard deviation in advantage normalization also helps mitigate reward hacking. In image generation tasks, reward models often assign overly similar rewards $R(x^i_0, c)$ to comparable images within the same group, causing an extremely small standard deviation, i.e., ${\rm Std}\to0$. This dangerously amplifies small, and often meaningless, reward differences in Eq.~\eqref{eq:adv_norm}, making training sensitive to reward noise and leading to over-optimization. While a concurrent work, Pref-GRPO~\citep{wang2025pref}, proposes to mitigate this issue by using a preference model to convert rewards into pairwise win-rates, this method is computationally expensive and lacks generality, as it relies on exhaustive pairwise comparisons and is restricted to preference-based reward models. In contrast, we propose a simpler, more efficient, and general solution: directly removing standard deviation (${\rm Std}$) from advantage normalization~\citep{liu2025understanding,he2025random}. This imposes a natural constraint when rewards are similar, preventing harmful amplification.

\begin{figure*}[htbp]
    \centering
    \includegraphics[width=1\linewidth]{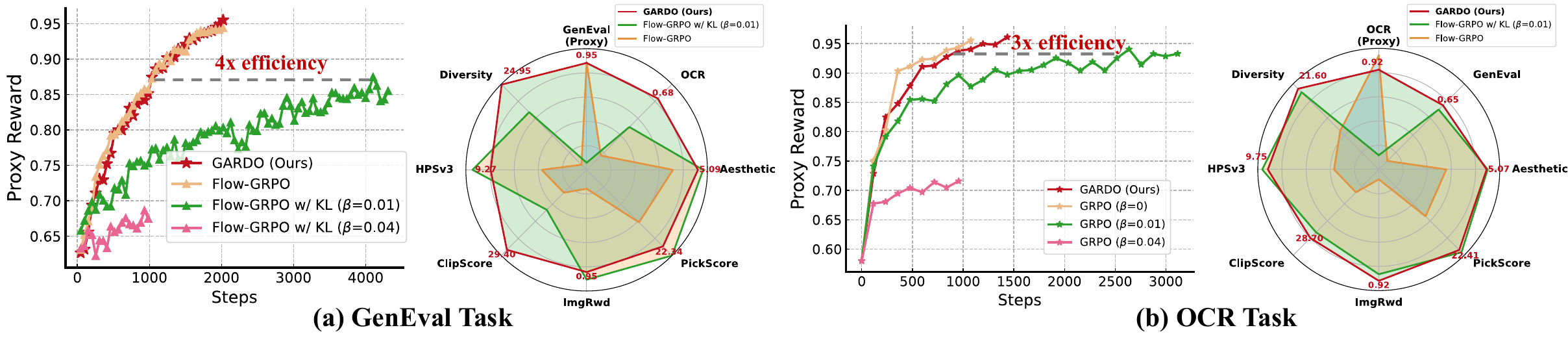}
    \vspace{-1.5em}
    \caption{Learning curves and o.o.d. generalization results across different methods. GARDO not only matches the sample efficiency of the KL-free baseline, but also mitigates reward hacking effectively, as evidenced by the superior performance on unseen metrics.}
    \label{fig:main_fig}
\end{figure*}

\section{Experiments}

\noindent\textbf{Setup.} Following Flow-GRPO~\citep{liu2025flowgrpo}, we choose SD3.5-Medium~\citep{esser2024scaling} as the base model and GRPO as the base RL algorithm for empirical validation. To demonstrate the versatility of our framework across different base models and RL algorithms, we also provide additional results on Flux.1-dev~\citep{flux2024} and on a distinctly different online RL algorithm, DiffusionNFT~\citep{zheng2025diffusionnft}, in Appendix~\ref{sec:results}. Throughout all experiments, images are generated at 512×512 resolution. We fine-tune the reference model with LoRA~\citep{hu2022lora} ($\alpha=64,r=32$). We set group size $G=24$ for estimating the diversity-aware advantages. 

\noindent\textbf{Benchmarks.}  We employ multiple tasks with diverse metrics to evaluate the performance of our method for preventing reward hacking without sacrificing sample efficiency. We employ GenEval~\citep{ghosh2023geneval} and Text Render (OCR)~\citep{chen2023textdiffuser} as the proxy tasks. GenEval evaluates the model's generation ability on complex compositional
prompts, including 6 different dimensions like object counting, spatial relations, and attribute binding. OCR measures the text accuracy of generated images. For these two tasks, we use the corresponding training and test sets from Flow-GRPO. We employ unseen metrics including Aesthetic~\citep{schuhmann2022laion}, PickScore~\citep{kirstain2023pick}, ImageReward~\citep{xu2023imagereward}, ClipScore~\citep{hessel2021clipscore}, and HPSv3~\citep{Ma2025HPSv3TW} for assessing the o.o.d. generalization performance. For evaluating the diversity of the generated images, we use the mean of pairwise cosine distance across a group of images for quantification, i.e., ${\rm Div}={\rm mean}_{i,j\in[1,G],i\neq j}(1-\frac{e_i\cdot e_j}{|e_i||e_j|})$. More implementation details are provided in Appendix~\ref{sec:imp_details}.
\subsection{Results Analysis}
\noindent\textbf{Removing Std from Advantage Normalization is Useful.} As shown in Table~\ref{table:main_result}, eliminating standard deviation normalization alleviates reward hacking and improves performance on unseen rewards compared to the baseline, while largely preserving sample efficiency and high proxy rewards. However, its performance on unseen metrics still falls short of the reference model. This observation clearly demonstrates that, while this technique is valuable, it is insufficient on its own to fully resolve the reward hacking problem as defined in Definition~\ref{def:hack}.

\begin{figure*}[tbp]
\centering\includegraphics[width=1.\linewidth]{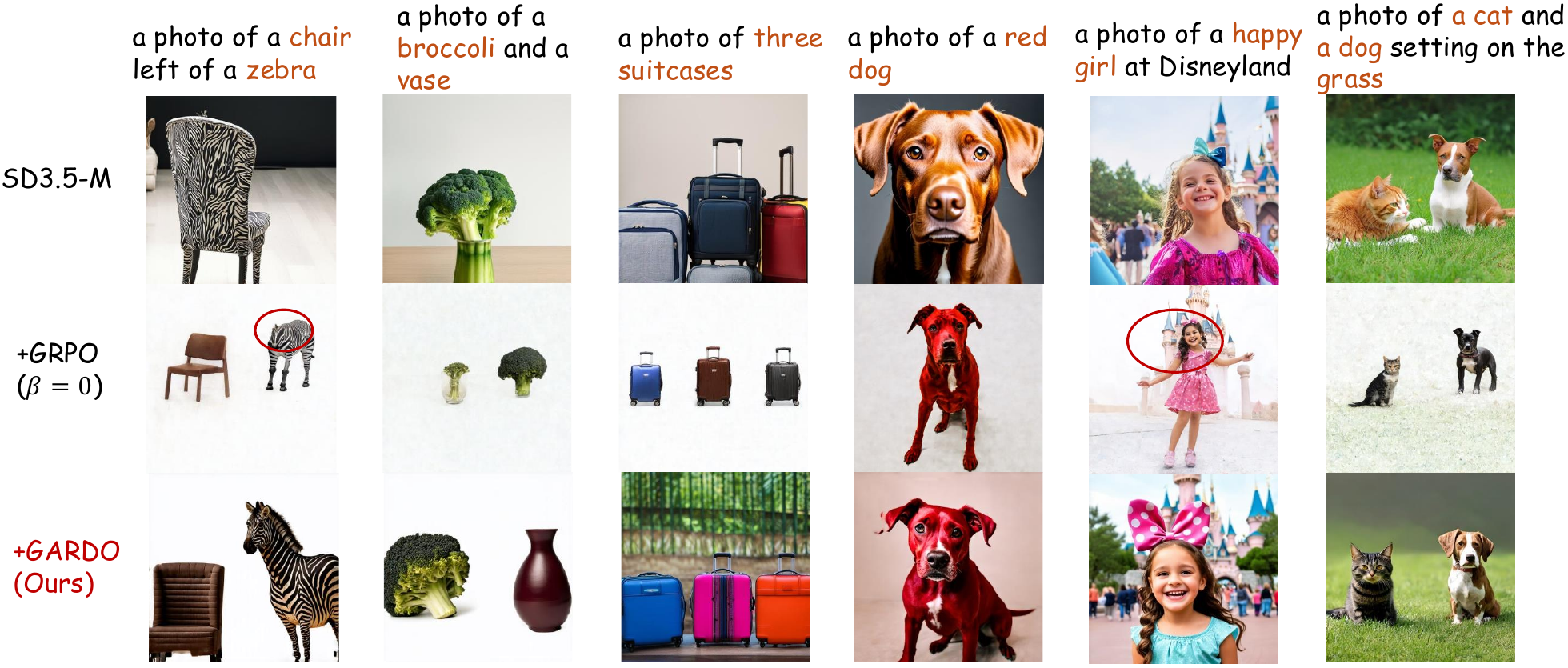}
    \caption{Qualitative images generated by GARDO and vanilla GRPO across different prompts. Only GARDO generates images correctly aligned with the prompt, while maintaining a satisfactory perceptual quality and diversity.}
    \label{fig:demo-hacking}
    \vspace{-.8em}
\end{figure*}

\noindent\textbf{Efficiency vs. Reward Hacking.} The results in Table~\ref{table:main_result} highlight a critical trade-off inherent in RL-based fine-tuning. The GRPO ($\beta=0$) baseline, while achieving a high proxy reward, suffers from severe reward hacking, as evidenced by its poor performance on unseen metrics such as Aesthectic, HPSv3, and diversity. Adding a KL penalty mitigates this over-optimization but at a significant cost to sample efficiency, resulting in a proxy reward that is over 10 points lower given the same computational budget. In contrast, our proposed method successfully reconciles this trade-off. As illustrated in Fig.~\ref{fig:main_fig}, it achieves a proxy reward comparable to the strongest KL-free baseline while simultaneously preserving high unseen rewards. Notably, GARDO's generalization performance not only matches but in some cases surpasses that of the original reference model. Quantitative results on GenEval tasks are provided in Fig.~\ref{fig:demo-hacking}. After training over the same steps, we find that only GARDO successfully follows the instruction and generates high-quality images, while vanilla GRPO obviously hacks the proxy reward, generating noisy images with blurred backgrounds and Gibbs artifacts.
\begin{figure}[htbp]
    \centering
    \includegraphics[width=1\linewidth]{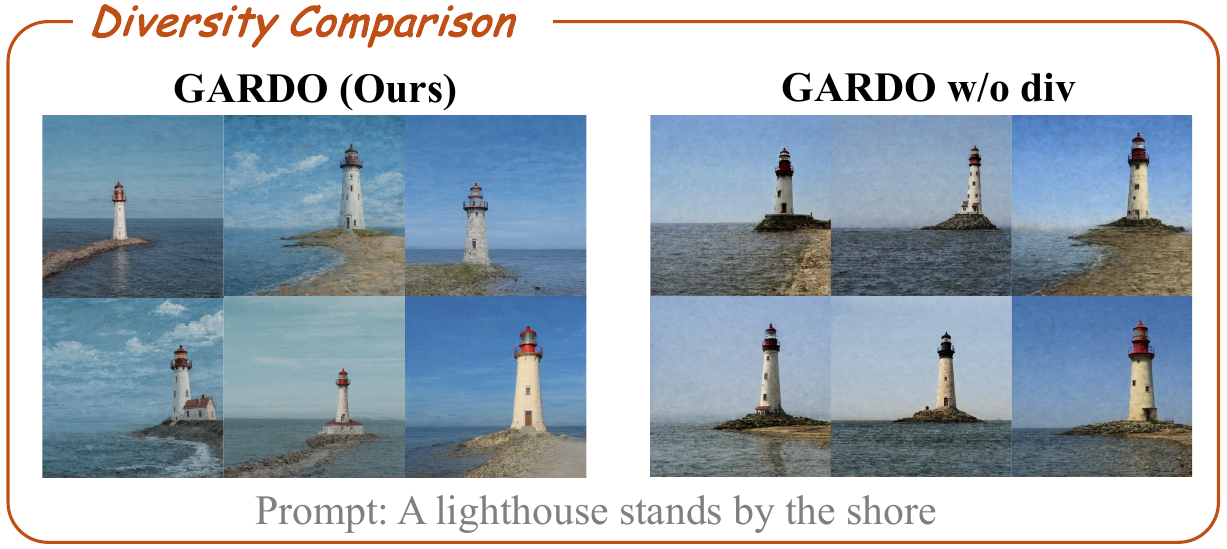}
    \vspace{-1em}
    \caption{Our diversity-aware advantage shaping effectively improves the generation diversity.}
    \vspace{-1em}
    \label{fig:demo_div}
\end{figure}

\noindent\textbf{Diversity-Aware Advantage Shaping Improves Diversity.} The results in Table~\ref{table:main_result} demonstrate that our proposed diversity-aware optimization leads to a remarkable increase in sample diversity scores, i.e., $19.98\to24.95$ for GenEval. Visualization is provided in Fig.~\ref{fig:demo_div} to further validate this efficacy. This increased diversity broadens the policy's exploration space, enabling it to discover novel states beyond the initial data distribution. This, in turn, prevents convergence to a narrow set of solutions (i.e., mode collapse). Ultimately, the enhanced exploration translates into improved final performance on both proxy and unseen metrics.

\begin{figure}[htbp]
    \centering
    \vspace{-1.em}
    \includegraphics[width=.96\linewidth]{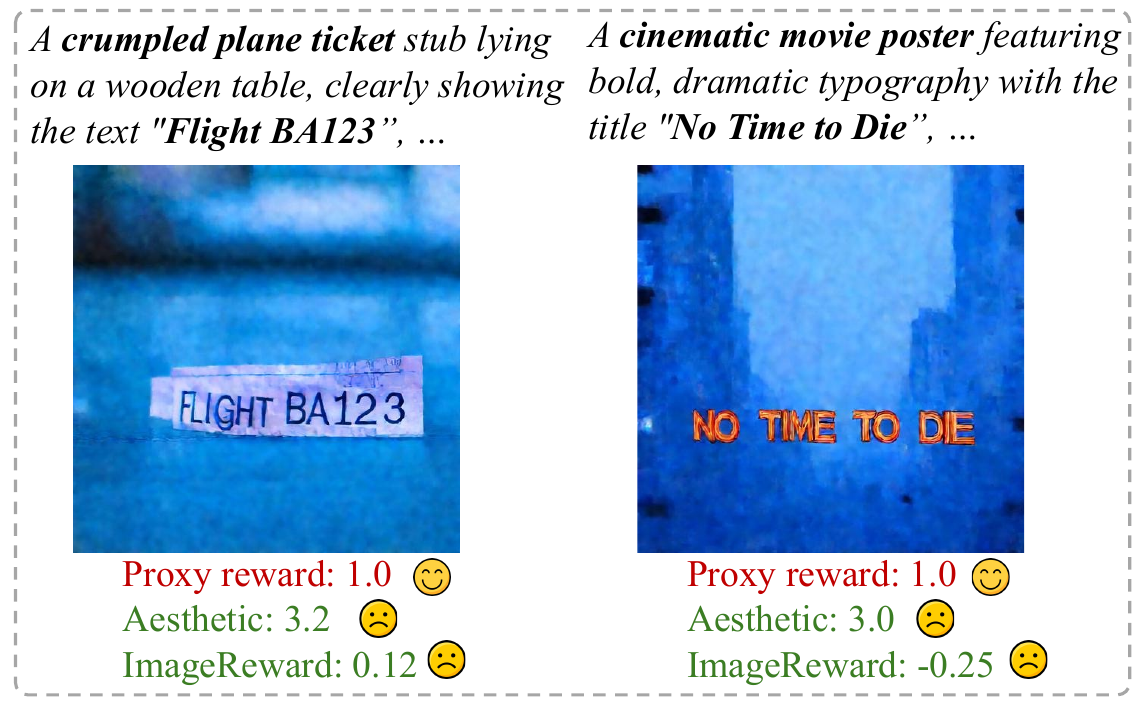}
    \vspace{-1.em}
    \caption{Samples that are identified with high uncertainty over the training process. While these samples reach high proxy rewards, they get very low unseen rewards.}
    \label{fig:kl_dynamic}
    \vspace{-1em}
\end{figure}

\begin{figure}[htbp]
    \centering
    \subfloat[KL dynamics]{\includegraphics[width=0.52\linewidth]{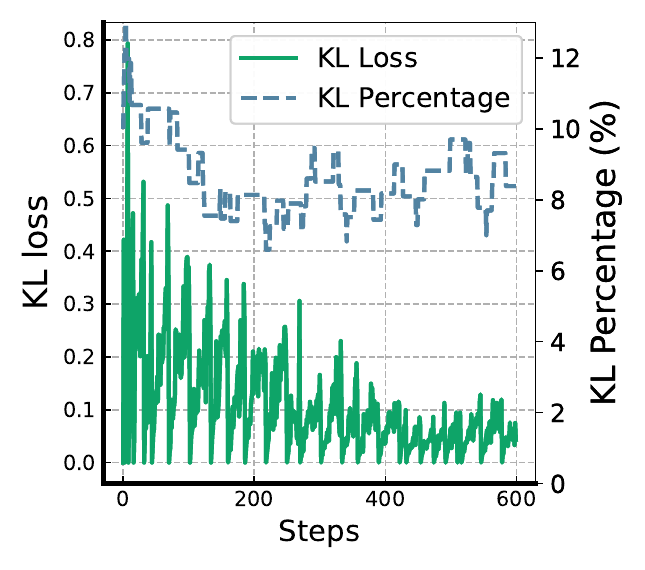}\label{fig:kl_dynamic_curve}}
     \subfloat[Compare w/ multiple rewards]{\includegraphics[width=0.48\linewidth]{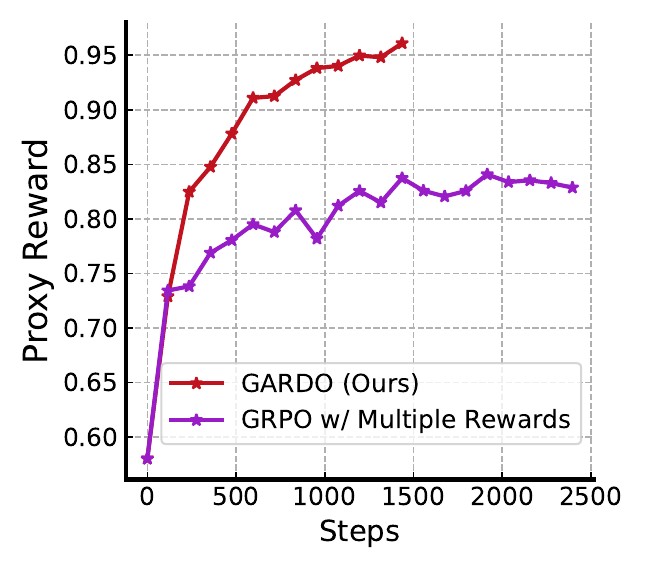}\label{fig:multi_rew}}
    \caption{(a): Dynamics of both KL loss and gated KL percentage throughout training. (b): Reward curves of GARDO and multi-reward GRPO on the OCR task.}
    \label{fig:kl_multi}
\end{figure}

\noindent\textbf{Gated and Adaptive KL Enhances Sample Efficiency without Reward Hacking.} As shown in Table~\ref{table:main_result}, `GARDO w/o div', which comprises only the gated and adaptive KL regularization, is sufficient to overcome the sample efficiency bottleneck of standard regularization, matching the convergence speed of the KL-free baseline while still mitigating reward hacking. It achieves a 0.91 OCR score given 600 steps without compromising the unseen rewards. This success is accomplished through two key components: (1) The adaptive regularization dynamically updates the reference model to prevent excessive regularization from a sub-optimal anchor, and (2) the gated KL identifies and applies a relatively high regularization specifically to ``illusory" samples with high reward uncertainty, thus avoiding unnecessary penalties. The dynamics of KL percentage and KL loss are provided in Fig.~\ref{fig:kl_dynamic_curve}, where only around 10\% of the samples are penalized. Fig.~\ref{fig:kl_dynamic} shows the examples that are identified as highly uncertain and thus penalized.

\noindent\textbf{Comparison with Multiple Reward Training.} While we leverage an ensemble of off-the-shelf reward models for uncertainty estimation, the policy itself is optimized against only a single proxy reward. This differentiates our approach from multi-objective reinforcement learning (MORL) methods, which seek to balance a weighted combination of multiple, often competing, reward signals. To demonstrate the superiority of our approach, we compare GARDO with the policy trained by multiple rewards, i.e., $0.8\times {\rm OCR (Proxy)}+0.1\times {\rm Aesthetic}+0.1\times {\rm Imagereward}$. This baseline is unregularized to maximize the sample efficiency. From the results shown in Fig.~\ref{fig:multi_rew}, we observe that RL trained with multiple rewards exhibits significantly lower sample efficiency with respect to the primary OCR proxy reward. This finding is aligned with previous multi-objective RL literature~\citep{yang2019generalized}, highlighting the challenges of optimizing for conflicting or misaligned rewards.

\subsection{Emerging Behavior}

\begin{figure}[htbp]

    \centering
    \includegraphics[width=.9\linewidth]{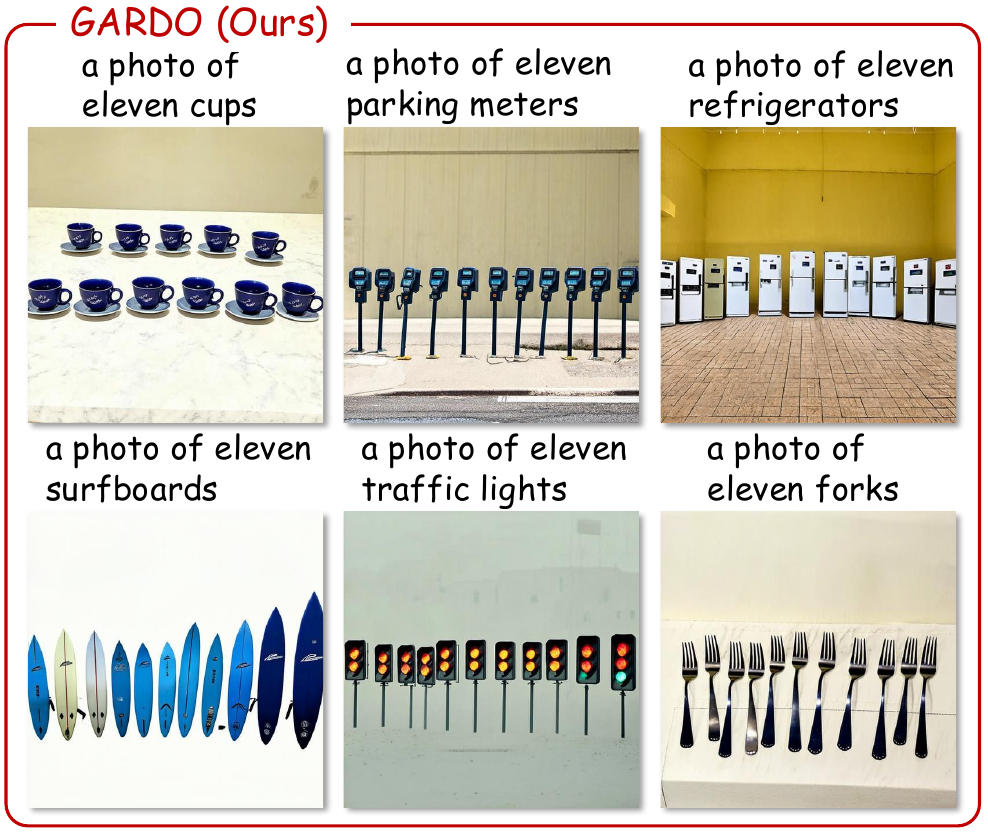}
    
    \caption{Visualization of GARDO counting 11 objects.}
    \vspace{-1em}
    \label{fig:count-demo}
\end{figure}

By eliminating universal KL regularization for all samples and periodically resetting the reference model according to the learning dynamic, along with the diversity-aware optimization, we can effectively unlock the emerging behavior that is missing from the base model~\citep{guo2025deepseek}. We validate this using a challenging counting task: the model is trained on datasets containing 1-9 objects and then tested on its ability to generate 10-11 objects, where the base model consistently fails. The results in Table~\ref{tab:counting} show that GARDO significantly improves the counting accuracy. This is particularly evident in the difficult task of generating 10 objects, where the base model exhibits near-zero accuracy. We provide the visualization results of GARDO counting 11 objects in Fig.~\ref{fig:count-demo}.
\begin{table}[!h]

\caption{Counting accuracy across vanilla GRPO with different KL coefficient and our method.}

    \label{tab:counting}
    \centering
    \resizebox{\linewidth}{!}{
    \begin{tabular}{ccccc}
    \toprule
    \multirow{2}{*}{\bf Method}& \multirow{2}{*}{\bf \#Step} & \multirow{2}{*}{\bf Trained Tasks}& \multicolumn{2}{c}{\bf Unseen Tasks}\\\cmidrule(lr){4-5}
 &&&Counting 10& Counting 11\\ 
\hline
\textcolor{gray}{SD3.5-M~\cite{rombach2022high}}  & - & \textcolor{gray}{0.28} & \textcolor{gray}{0.01} & \textcolor{gray}{0.01} \\\hline
      GRPO ($\beta=0.04$)  &2000 &0.41&0.09&0.07 \\
      GRPO ($\beta=0.01$)  &2000 & 0.56&0.27 &0.15 \\
      GRPO ($\beta=0$)  &2000 & 0.77&0.28 &0.15 \\\hline
      GARDO (Ours) &2000& \textbf{0.77} &\textbf{0.38} &\textbf{0.18}\\\bottomrule
    \end{tabular}
    }
   
\end{table}

\section{Conclusion and Limitations}
In this paper, we propose a novel and effective approach to address the challenge of reward hacking in fine-tuning diffusion models. We introduce a gated and adaptive regularization mechanism for more fine-grained control, and a diversity-aware strategy to encourage mode coverage, which significantly enhances sample efficiency and emerging behaviors. Nevertheless, a primary limitation of our method is its dependency on auxiliary reward models for uncertainty estimation. Consequently, the scalability of our approach to resource-intensive video generative models remains an open question for future investigation.
{
    \small
    \bibliographystyle{ieeenat_fullname}
    \bibliography{main}
}
\clearpage
\setcounter{section}{0}
\renewcommand{\thesection}{\Alph{section}}

\section{Implementation Details}
\label{sec:imp_details}
\begin{table*}[ht]
\caption{Results on \textcolor{orange}{GenEval} tasks for DiffusionNFT algorithm. GARDO successfully surpasses baselines in terms of both sample efficiency in proxy reward and generalization on o.o.d. rewards.
}
\vspace{-2.em}
\label{table:diffusionnft}
\begin{center}
\resizebox{\linewidth}{!}{
\begin{tabular}{ccllllllll}
\toprule
 \multirow{2}{*}{\bf Method}& \multirow{2}{*}{\bf \#Step} & \multicolumn{2}{c}{\bf Trained Tasks}& \multicolumn{6}{c}{\bf Unseen Tasks}\\\cmidrule(lr){3-4} \cmidrule(r){5-10}
 &&\multirow{1}{*}{\bf GenEval}  & \multirow{1}{*}{\bf OCR} & \multicolumn{1}{l}{\bf Aesthetic} & \multicolumn{1}{l}{\bf PickScore} &\multicolumn{1}{l}{\bf ImgRwd} & \multicolumn{1}{l}{\bf ClipScore} &\multicolumn{1}{l}{\bf HPSv3} &\multicolumn{1}{l}{\bf Diversity}\\ 
\hline
\textcolor{gray}{SD3.5-M~\cite{rombach2022high}}  & - & \textcolor{gray}{0.63} & \textcolor{gray}{0.58} & \textcolor{gray}{5.07} & \textcolor{gray}{22.40}  & \textcolor{gray}{0.83} & \textcolor{gray}{28.2}& \textcolor{gray}{9.70}& \textcolor{gray}{21.84}\\\hline

+DiffusionNFT ($\beta$=0)  & 400 & \cellcolor[rgb]{1.0,0.9,0.8}\underline{0.94} &0.68 &4.23  & 21.85 & 0.62 & 28.8&5.66&11.78\\
+DiffusionNFT ($\beta$=0.04)  &  400    &\cellcolor[rgb]{1.0,0.9,0.8}0.72  & 0.54 & 4.87 & 22.54 & 1.14 & 29.1&11.51&13.87\\
+DiffusionNFT ($\beta$=0.04)  &  1200    &\cellcolor[rgb]{1.0,0.9,0.8}0.90  & 0.51 & 4.55 & 21.96 & 0.89 & 29.0&10.04&13.70\\
\hline
\rowcolor[rgb]{0.9,1.0,0.8}+GARDO (Ours) ($\beta$=0.04) &400  &\cellcolor[rgb]{1.0,0.9,0.8}0.95 & \textbf{0.64}    &\textbf{5.23}  &\textbf{22.59}  & \textbf{1.10} &\textbf{29.2}  & \underline{12.17}&\textbf{14.57}\\

\bottomrule
\end{tabular}
}
\end{center}
\vspace{-2.2em}
\end{table*}
\subsection{Details of GARDO}
We choose Flow-GRPO~\citep{liu2025flowgrpo} as the base RL algorithm to implement GARDO. We provide the training details and related hyperparameters as follows:
\begin{itemize}
    \item Following Flow-GRPO~\citep{liu2025flowgrpo}, we set the sampling timestep $T = 10$ and an evaluation timestep $T = 40$.
    \item We set the training batch size as 6, and set the group size $G=24$.
    \item Learning rate is $3e^{-4}$, and the clip range is $1e^{-4}$.
    \item We use Lora with $\alpha=32$ and $r=64$.
    \item Given a group of images generated with the same prompt, the initial noises are set to be the same to ensure GRPO's assumptions.
\end{itemize}
\paragraph{GARDO's adaptive KL.} We set the KL threshold $\epsilon_{\rm KL}=1e-4$. The maximum gradient steps $m$ between two consecutive reset operations is set to be $100$. This means reset happens if the KL loss surpasses $\epsilon_{\rm KL}$ or there have been $m$ gradient steps after the last recent reset.

\paragraph{GARDO's gated KL} We set the initial KL percentage $k=0.1$. This means the top 10\% samples with the highest uncertainty $\mathcal{U}$ are penalized at the beginning of the training stage. $k$ is dynamically updated throughout the training process. We maintain a cache window to determine $k$. The window size is 20, which stores the maximum and minimum value of uncertainty $\mathcal{U}$ over 20 epochs. If the mean value of the uncertainty within the current batch is higher than the maximum value cached in the window, then we set $k\gets k\times 1.1$; if the mean value of the uncertainty within the current batch is lower than the minimum value cached in the window, $k\gets k\times0.9$; otherwise, $k$ remains unchanged.

We summarize our method in the pseudocode in Alg.~\ref{alg:gardo}.
\begin{algorithm}[htbp]
\caption{Overview of GARDO}
\label{alg:gardo}
\begin{algorithmic}[1]
\Require initial policy model $\pi_\theta$; proxy reward $\tilde{R}$; auxiliary reward models $\hat{R}_1,\hat{R}_2$; Dinov3 model $f_\phi$; KL loss threshold $\epsilon_{\rm KL}$; maximum reset steps $m$; prompt dataset $\mathcal{C}$; total sampling steps $\tilde{T}$; number of samples per prompt $G$; Gated KL percentage $k$.
\State Set reference policy $\pi_{\rm ref}\gets\pi_\theta$
\State Uncertainty window $\mathcal{W}_{\mathcal{U}} \gets \emptyset$ with a window size $W$
\For{training iteration $n=1$ \textbf{to} $N$}
    \State Sample batch prompts $\mathcal{C}_b \sim \mathcal{C}$
    \State Update old policy model: $\pi_{\theta_{\text{old}}} \gets \pi_\theta$
    \State $\mathcal{L}_{total}\gets0,\bar{\mathcal{U}}_{batch}\gets 0 $
    \For{each prompt $\mathbf{c} \in \mathcal{C}_b$}
        \State Initial the same noise $\epsilon \sim \mathcal{N}(0,\mathbf{I})$
        \State Generate G images $\{x_0\}_{i=1}^G$ after $\tilde{T}$ sampling steps.
        \State Obtains advantages $A_{i=1}^G$ from $\tilde{R}$ via Eq.~\ref{eq:adv_norm}.
        \State Reshape $A_{i=1}^G$ and get diversity-aware advantages by Eq.~\ref{eq:diversity}.
        \State Obtains uncertainty estimation $\mathcal{U}^G_{i=1}$ via Eq.~\ref{eq:uncertainty}.
        \State Determine gate threshold: $\epsilon_\mathcal{U} = \text{Percentile}(\{\mathcal{U}^i\}_{i=1}^G, 1-k)$.
        \State Calculate KL-regularized loss $\mathcal{L}_{total}\gets\mathcal{L}_{total}+\frac{1}{G} \sum_{i=1}^G [ \mathcal{L}_{\rm RL}^i + \mathbb{I}(\mathcal{U}^i > \epsilon_\mathcal{U}) \cdot \mathcal{L}_{\rm KL}^i ]$
        \State $\bar{\mathcal{U}}_{batch} \gets \bar{\mathcal{U}}_{batch} +(\text{mean}(\{\mathcal{U}^i\}))$
      
        \EndFor
        
        \State Update policy model via gradient ascent: $\theta \gets \theta + \eta\nabla_\theta\mathcal{L}_{total}$
        \If{$n\ {\rm mod}\ m\ {\rm is}\ 0\ {\rm or}\ \mathcal{L}_{\rm KL}>\epsilon_{\rm KL}$}
        \State Reset $\pi_{\rm ref}\gets\pi_\theta$
        \EndIf
        \If{$|\mathcal{W}_\mathcal{U}| \ge W$}
        \If{$\bar{\mathcal{U}}_{batch} > \max(\mathcal{W}_\mathcal{U})$} 
        \State $k \gets \min(k \times 1.1, 1.0)$
        \EndIf
        \If{$\bar{\mathcal{U}}_{batch} < \min(\mathcal{W}_\mathcal{U})$} 
        \State $k \gets k \times 0.9$
        \EndIf
    \EndIf
    \State Update window $\mathcal{W}_\mathcal{U}$: Append $\bar{\mathcal{U}}_{batch}$, remove oldest if $|\mathcal{W}_\mathcal{U}|> W$.

    \EndFor

\end{algorithmic}
\end{algorithm}

\subsection{Computation  Specification}
We train our model using 8 NVIDIA A800 GPUs. It usually takes around 49s for a gradient step, including both sampling and training. 
\subsection{Evaluation Metrics}
\paragraph{Proxy Tasks.}
In this paper, we include two proxy tasks to train the diffusion models by RL, i.e., GenEval and OCR tasks. We ensure the training hyperparameters are the same across different methods for fair comparison. Below, we introduce the proxy tasks used in the paper.
\begin{itemize}
    \item The GenEval framework following Flow-GRPO’s
experimental protocol. The training dataset is sourced from the Flow-GRPO dataset. GenEval includes six difficult compositional image generation tasks. We use its official evaluation pipeline, which detects object co-occurrence, spatial positioning, object count, and color attributes for fine-grained assessment.
    \item For the OCR task, we use the training dataset and test dataset from Flow-GRPO. This task measures text rendering accuracy with the reward $r = \max(1-Ne/N_{ref},0)$, where $Ne$ is the minimum edit distance between the rendered text and the target text and $N_{ref}$ is the number of characters required to render.
\end{itemize}
\paragraph{Unseen Tasks}
We adopt DrawBench~\citep{saharia2022photorealistic} for evaluation, which consists of 200 prompts spanning 11 different categories, serving as an effective test set for comprehensive evaluation of the T2I models. For each prompt, we generate four images for evaluation for a fair and convincing comparison. We employ Aesthetic~\citep{schuhmann2022laion}, PickScore~\citep{kirstain2023pick}, ImageReward~\citep{xu2023imagereward}, ClipScore~\citep{hessel2021clipscore}, and HPSv3~\citep{Ma2025HPSv3TW} for extensive evaluation of the o.o.d. generalization ability, and detecting the degree of reward hacking. For the diversity score, we employ Dinov3~\citep{simeoni2025dinov3} to extract feature embeddings $e^i$ of each image. Then we use the mean of pairwise cosine distance across a group of images for diversity quantification:
\begin{equation}
    {\rm Div}={\rm mean}_{i,j\in[1,G],i\neq j}(1-\frac{e_i\cdot e_j}{|e_i||e_j|}),
\end{equation}
where a group of images is generated, given the same prompt.
\section{Additional Results}
\label{sec:results}
\subsection{Results on DiffusionNFT~\citep{zheng2025diffusionnft}}
GARDO is built upon an existing regularized-RL objective, which can be compatible with various RL algorithms. To demonstrate GARDO's versatility, we apply it to a recently released RL algorithm, DiffusionNFT~\citep{zheng2025diffusionnft}, which can be very different from GRPO. DiffusionNFT directly optimizes velocity without relying on the computation of log likelihood, bridging the gap between SFT pre-training and RL post-training. From the results shown in Table~\ref{table:diffusionnft}, we have the following observations: (1) Similar to Flow-GRPO, vanilla DiffusionNFT still suffers from reward hacking, as evidenced by the reduced performance on unseen tasks like Aesthetic, picksocre, and HPSv3. (2) Adding KL regularization can effectively mitigate reward hacking, enabling more robust optimization. However, it significantly compromises the sample efficiency. Given the same training steps (i.e., 400 steps), KL-regularized DiffusionNFT only achieves a 0.72 accuracy on the proxy task, i.e., GenEval. (3) GARDO performs best in a balance between sample efficiency and the mitigation of reward hacking. GARDO can obtain the highest score (i.e., 0.95) on GenEval given 400 steps without reducing unseen rewards compared to the reference model. It even boosts the unseen reward, like Aesthetic, and remains the highest diversity compared with baselines. This clearly demonstrates that GARDO achieves the highest sample efficiency while effectively preventing hacking, showing superior performance in terms of both proxy rewards and unseen rewards.

\subsection{Results on Flux.1-dev}
\begin{figure}[htbp]
    \centering
    \vspace{-1.em}
    \subfloat[]{\includegraphics[width=0.43\linewidth]{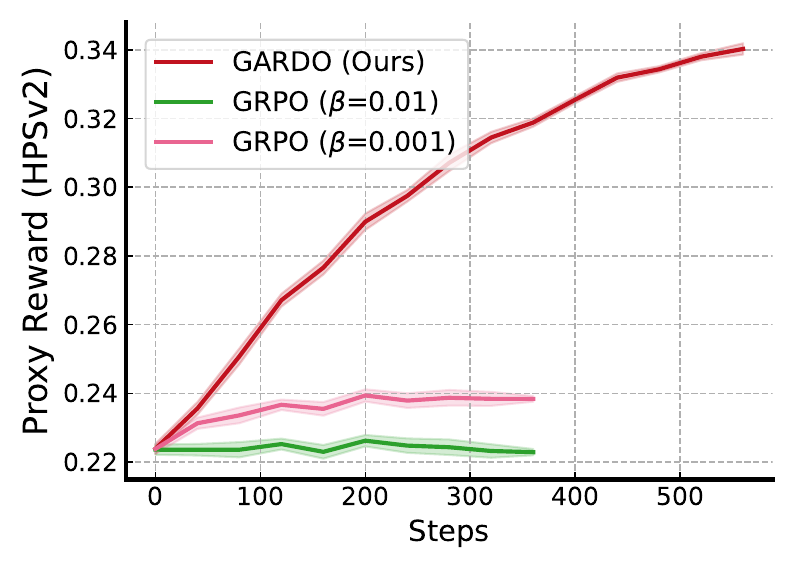}
   
    \label{fig:flux_hpsv2}}
    \subfloat[]{\includegraphics[width=0.56\linewidth]{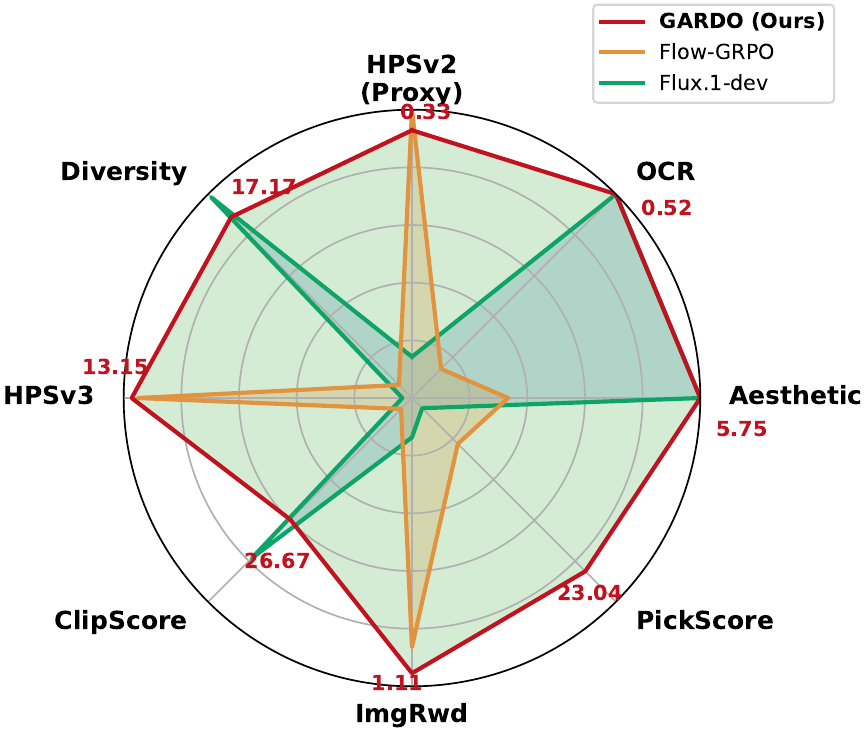}
    
    \label{fig:flux-radar}}
    \vspace{-1em}
    \caption{(a): Learning curves of GARDO on Flux.1-dev with HPSv2 as the proxy reward. (b): Results on both proxy reward and o.o.d. unseen rewards. GARDO performs best in the trade-off.}
    \vspace{-1.5em}
\end{figure}

\begin{figure*}[htbp]
    \centering
    \includegraphics[width=1\linewidth]{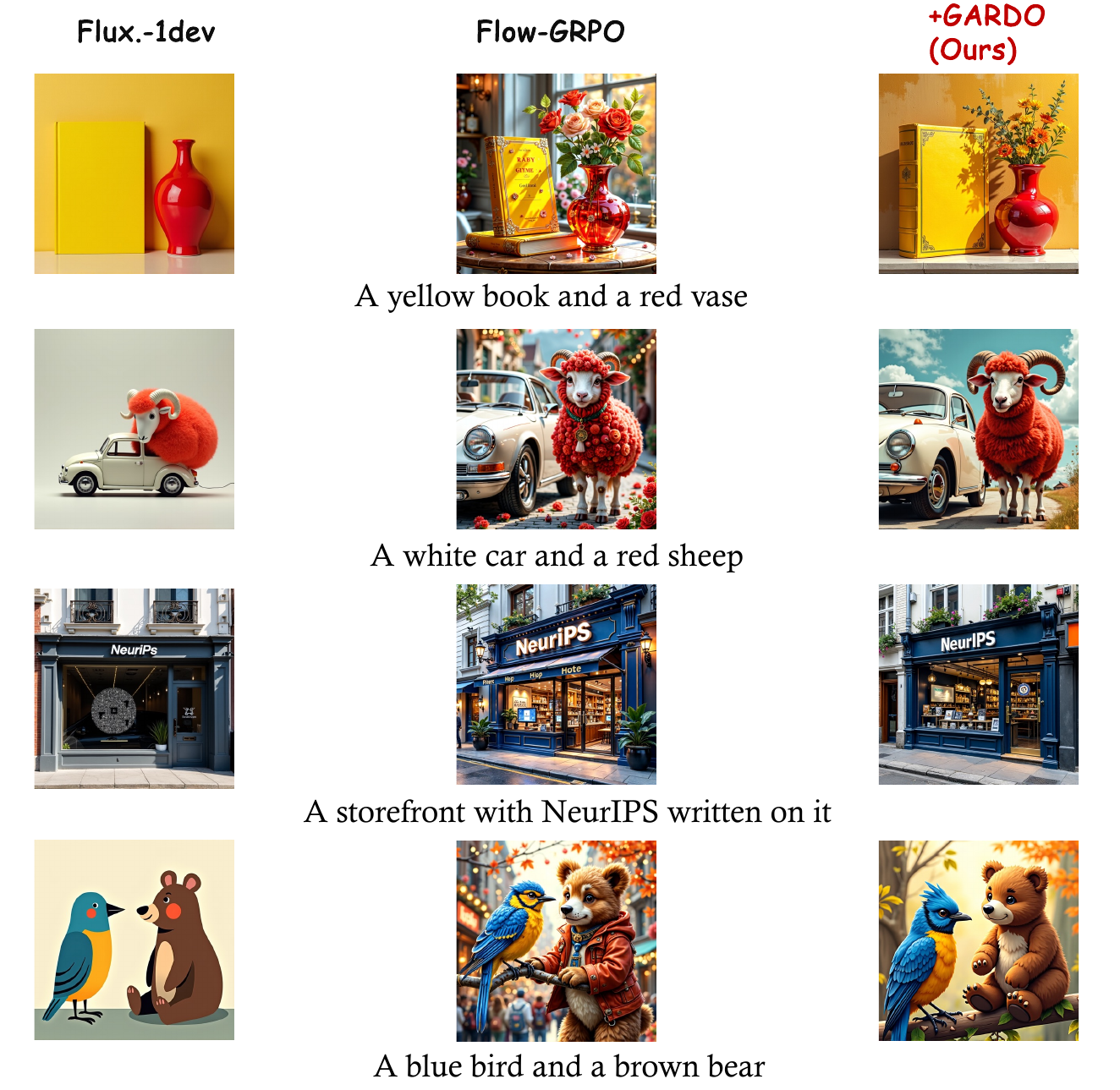}
    \caption{Qualitative results on Flux.1-dev across GARDO and baselines. \textcolor{red}{While both GRPO and GARDO significantly improve the visual quality, GRPO tends to hack the HPSv2 reward, generating unnecessary or even undesired details.}}
    \label{fig:flux-demo}
\end{figure*}

\begin{figure*}[htbp]
    \centering
    \includegraphics[width=1\linewidth]{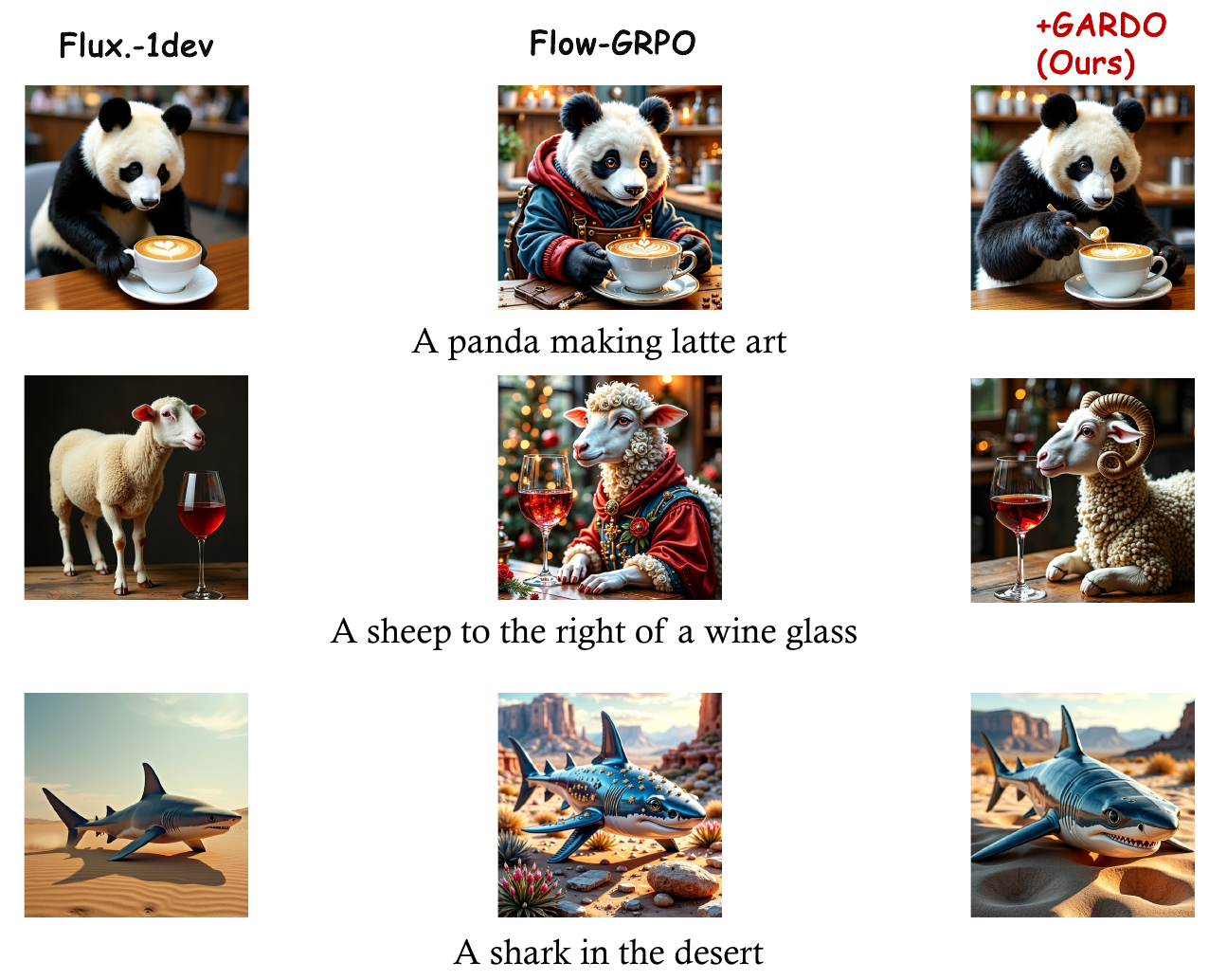}
    \caption{Qualitative results on Flux.1-dev across GARDO and baselines. \textcolor{red}{While both GRPO and GARDO significantly improve the visual quality, GRPO tends to hack the HPSv2 reward, generating unnecessary or even undesired details.}}
    \label{fig:flux-demo-1}
\end{figure*}

GARDO can also generalize to different base models. We choose Flux.1-dev~\citep{flux2024} as the base model, which contains 12B parameters and is known as the SOTA model for text-to-image generation. Here, we leverage HPSv2 as the proxy reward for optimization, using open-sourced HPDv3 (\url{https://huggingface.co/datasets/MizzenAI/HPDv3}) as the training dataset. GARDO still achieves the highest sample efficiency compared with the KL-regularized Flow-GRPO method. We provide the reward curves in Fig.~\ref{fig:flux_hpsv2} and qualitative demos in Fig.~\ref{fig:flux-demo} and Fig.~\ref{fig:flux-demo-1}. As shown in Fig.~\ref{fig:flux-radar}, GARDO outperforms both Flow-GRPO and base models in generalizing to unseen rewards, indicating GARDO's great potential to optimize any proxy rewards without reward hacking.

\subsection{More Qualitative Results}
\begin{figure*}[htbp]
    \centering
    \includegraphics[width=1\linewidth]{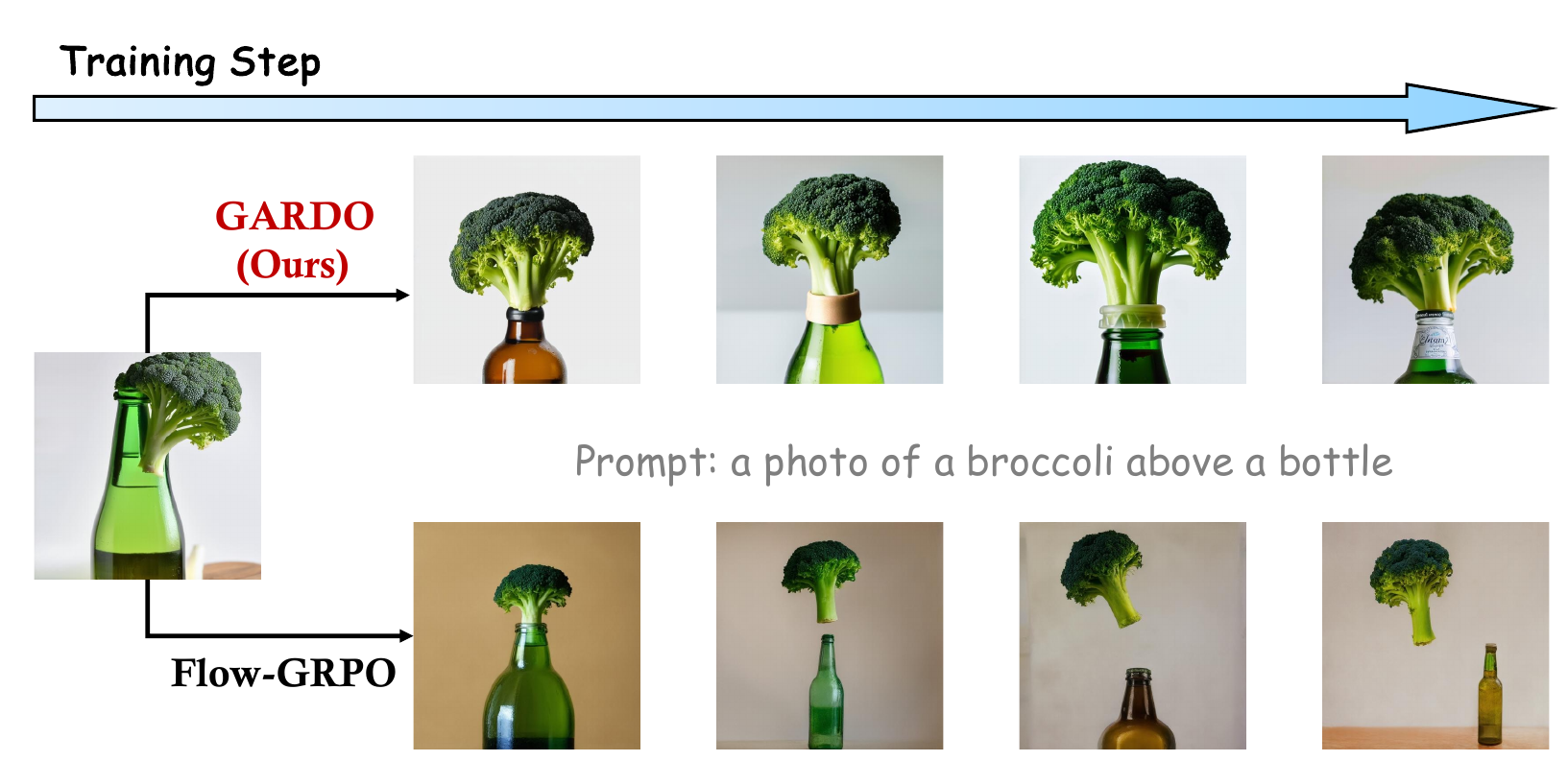}
    \caption{Generated images along the training process.}
    \label{fig:comp1}
\end{figure*}

\begin{figure*}[htbp]
    \centering
    \includegraphics[width=1\linewidth]{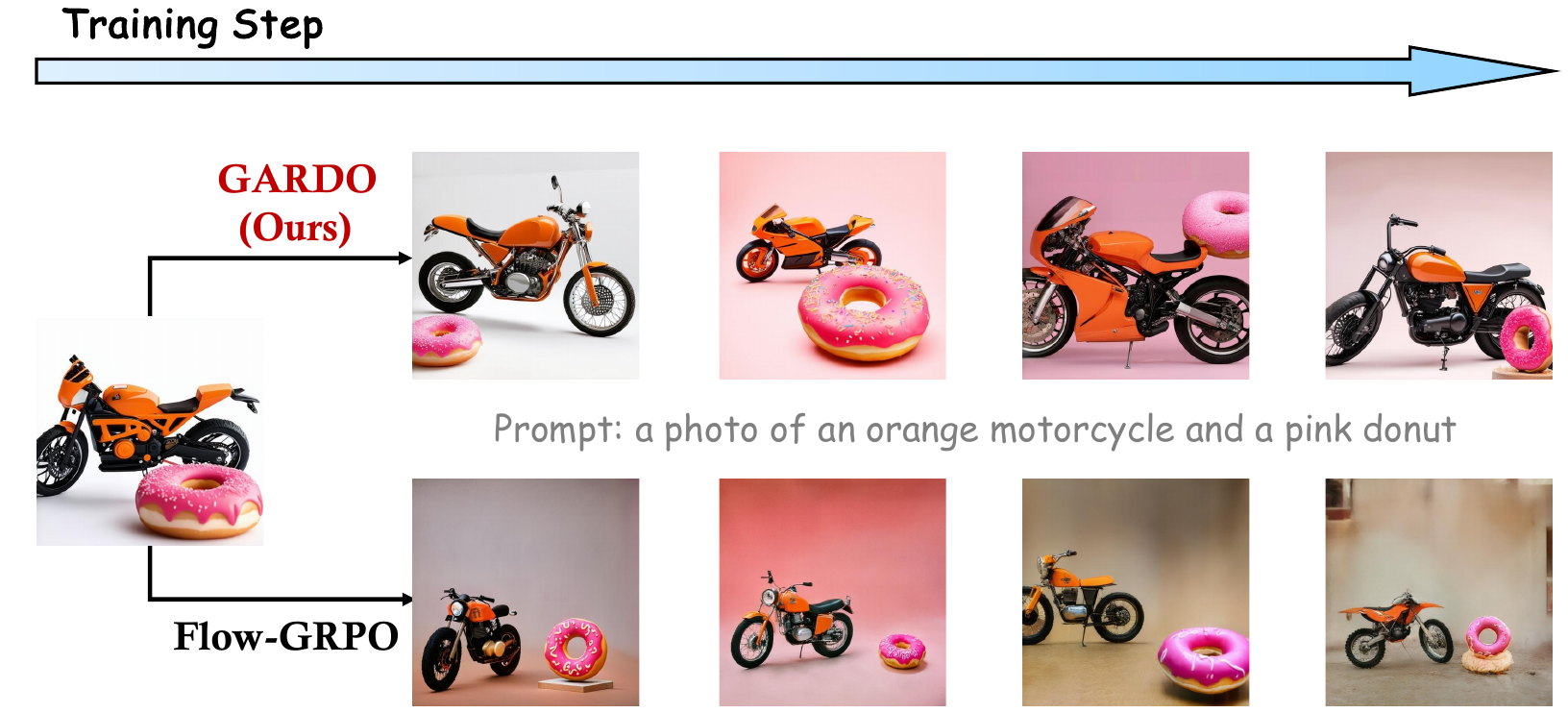}
    \caption{Generated images along the training process.}
    \label{fig:comp2}
\end{figure*}

We provide the generated images along the training process in Fig.~\ref{fig:comp1} and Fig.~\ref{fig:comp2}. As the training step increases, we observe that Flow-GRPO obviously hacks the reward ( or exploits the flaws), yielding reduced perceptual visual quality. However, GARDO remains a high visual quality throughout the training process, without compromising optimization performance on the proxy reward.

\end{document}